\documentclass[11pt,a4paper]{article}
\usepackage{times,latexsym}
\usepackage{url}
\usepackage[T1]{fontenc}

\usepackage[acceptedWithA]{tacl2021v1}

\usepackage{xspace,mfirstuc,tabulary}

\newif\iftaclinstructions
\taclinstructionsfalse 
\iftaclinstructions
\renewcommand{\confidential}{}
\renewcommand{\anonsubtext}{(No author info supplied here, for consistency with
TACL-submission anonymization requirements)}
\newcommand{\instr}
\fi

\iftaclpubformat 

\else

\fi

\usepackage{multirow}
\usepackage{graphicx}
\usepackage{array}             
\usepackage{rotating}
\usepackage{enumitem}
\usepackage{xcolor}  
\usepackage{booktabs}
\usepackage{soul}    

\title{Benchmarking the Generation of Fact Checking Explanations}

\author{
  Daniel Russo$^{1,2}$,
  Serra Sinem Tekiro\u{g}lu$^1$,
  Marco Guerini$^1$
  \\
  \ \\
  $^1$Fondazione Bruno Kessler, Via Sommarive 18, Povo, Trento, Italy
  \\
  \texttt{drusso@fbk.eu, tekiroglu@fbk.eu, guerini@fbk.eu}
  \\
  $^2$University of Trento, Italy
  \\
}

\date{}

\begin{document}
\maketitle

\begin{abstract}
Fighting misinformation is a challenging, yet crucial, task. Despite the growing number of experts being involved in manual fact-checking, this activity is time-consuming and cannot keep up with the ever-increasing amount of Fake News produced daily. Hence, automating this process is necessary to help curb misinformation. Thus far, researchers have mainly focused on claim veracity classification. In this paper, instead, we address the generation of justifications (textual explanation of \textit{why} a claim is classified as either true or false) and benchmark it with novel datasets and advanced baselines. In particular, we focus on summarization approaches over unstructured knowledge (i.e. news articles) and we experiment with several extractive and abstractive strategies. We employed two datasets with different styles and structures, in order to assess the generalizability of our findings. Results show that in justification production summarization benefits from the claim information, and, in particular, that a claim-driven extractive step improves abstractive summarization performances. Finally, we show that although cross-dataset experiments suffer from performance degradation, a unique model trained on a combination of the two datasets is able to retain style information in an efficient manner.
\end{abstract}

\section{Introduction}

The interaction between the modern media ecosystem and
online social media has facilitated the rapid and nearly unrestricted spreading of news. While this has been a major achievement in terms of access to information,
there is also an increasing need to counter the spread of misinformation, commonly conveyed through Fake News. Fake News is crafted with the intention to manipulate society towards a specific political, economic, or social outcome, lacking verifiable evidence and credible sources \citep{chen2015learning}. It can represent a threat to human health and safety, e.g. by disseminating false information on disease treatment \citep{van2022misinformation}. Thus, verifying the accuracy of claims and presenting users with factual and impartial evidence to support their veracity is of utmost importance. Manual fact-checking, however, is a time-consuming activity \citep{hassan2015quest}. Hence, Natural Language Processing has been suggested as an effective solution for automating this process. Thus far, the main strategies have involved classifying and flagging misleading information. However, a simple classification approach can generate a \textit{backfire effect} where the belief of false claims is further entrenched rather than hindered \citep{lewandowsky2012misinformation}. For this reason, explaining \textit{why} a claim is classified as either true or false can be a better solution. Fact-checking articles could represent a valuable resource towards this end, however, on online social media platforms they are ineffective either because ordinary users are not prone to click on links to relevant resources \citep{glenski2017consumers, glenski2020user} or because these articles are excessively long to the point that users would avoid reading it \citep{pernice2018people}. Indeed, effective explanations should be simple, and only a few arguments must be provided in order to avoid an \textit{``overkill'' backfire effect} \citep{lombrozo2007simplicity,sanna2006metacognitive}.

Although the work of professional fact-checkers is crucial for countering misinformation \citep{Wintersieck}, it has been shown that disproof on social media platforms is mostly carried out by ordinary users \citep{Micallef_et_al}. Thus, automating the explanation generation process is deemed crucial, as an aid for both fact-checkers (to increase their online activity) and for social media users \citep[to make their intervention more effective; ][]{he2023reinforcement}.

Still, few attempts to automatically generate explanations/justifications about claim veracity
have been proposed so far \cite{kotonya2020survey}. Current methods for justification production include highlighting tokens with high attention weights \citep{popat2018declare, yang2019xfake, lu2020gcan}, utilizing knowledge graphs \citep{ahmadi2019explainable}, and modelling it as either an extractive or abstractive summarization task \citep{atanasova2020generating,kotonya2020explainable}.

\begin{figure}[t!]{\includegraphics[width=1\columnwidth]{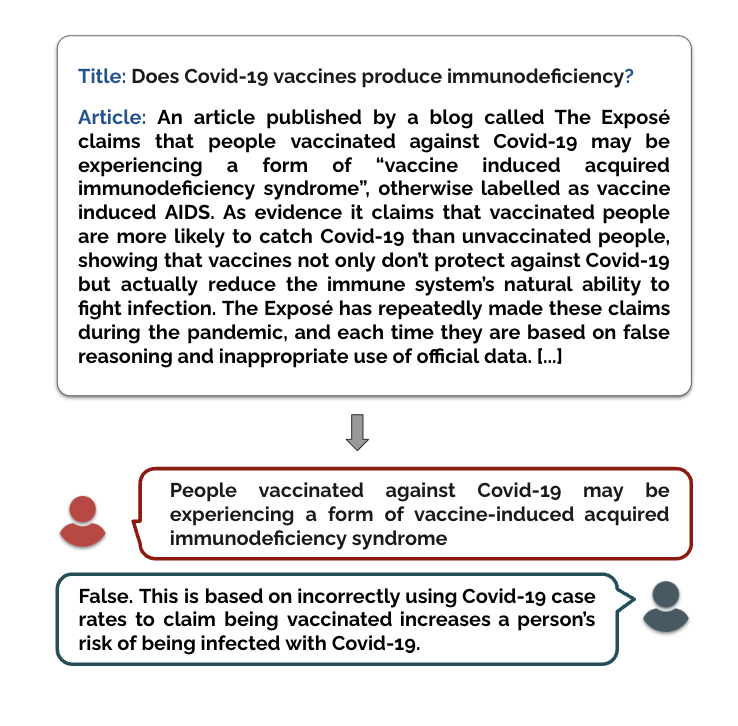}}
    \caption{An article (top) used to generate a verdict (bottom) in response to a false claim (middle).}
    \label{fig:dialogue_example}
 \end{figure}

In this paper, we aim at benchmarking justification production as a summarization task, by providing an exhaustive study of the performances of extractive and abstractive approaches over two novel datasets. In particular, we consider several extractive and abstractive approaches both in supervised and unsupervised settings, where we generate a justification for a given claim using a fact-checking article as a knowledge source. We also experiment with hybrid approaches combining extractive and abstractive steps in a unique pipeline. Finally, we integrate the pipeline within an end-to-end claim-driven explanation generation framework. These approaches are tested both in in-domain and cross-domain configurations, by employing two different datasets. Each dataset has its own style and characteristics, but they both contain claim, verdict, and article triplets (see Figure \ref{fig:dialogue_example}). 

The main findings from our experiments are:
(i) If an extractive approach is employed for justification production, then the sentence selection must be driven by the claim information. (ii) If no training data is available in cross-domain experiments, extractive approaches can be better than abstractive ones for justification production. (iii) High-quality justifications can be obtained by combining in a unique pipeline extractive and abstractive summarization approaches (using simple off-the-shelf LMs), and by driving  sentence selection and justification generation with the claim information.Still, differently from previous studies, we found that the sentences extracted from the article must retain their order rather than being rearranged according to some notion of relevance. (iv) LMs for abstractive summarization should be selected according to article length since there is not a one-fits-all solution: for shorter articles, 512 tokens input length LMs provide better results, while using models with 1024 input length is beneficial for longer examples. (v) Although cross-dataset experiments suffer from performance degradation, LM-based models are able to retain different verdict styles: fine-tuning a single LM on the union of datasets with different stylistic characteristics leads to performance similar to those obtained by fine-tuning a model for every single dataset. 

\section{Related Work}
The process of fact-checking a news story involves determining the truthfulness of a statement (\textit{Verdict Prediction}) and the generation of a written rationale for the verdict (\textit{Justification Production}).
The claim veracity is usually assessed through a classification task, both binary \citep{nakashole2014language,potthast2017stylometric, popat2018declare} and multi-class \cite{wang-2017-liar,thorne2018fever}, or through a multitask learning approach \cite{augenstein2019multifc}. Recently, researchers are focusing on developing datasets and systems for evidence-based Verdict Prediction. Among the most relevant dataset, notable examples include the FEVER dataset \citep{thorne2018fever}, SciFact \citep{wadden-etal-2020-fact}, COVID-fact \citep{saakyan-etal-2021-covid}, and PolitiHop \citep{ijcai2021p536}.

\textit{Justification Production} has proven to be more challenging than \textit{Verdict Prediction}. Several approaches have been suggested, including logic-based approaches \citep{gad2019exfakt, ahmadi2019explainable} or deep-learning and attention-based techniques \citep{popat2018declare, yang2019xfake, shu2019defend, lu2020gcan}. Nevertheless, casting justification production as a summarization task appears to be the most viable solution \citep{kotonya2020survey}. Thereby, explanations can be derived from manually written debunking articles either by selecting important sentences from the text \citep[\textit{extractive approach}; ][]{atanasova2020generating} or by generating a new one \citep[\textit{abstractive approach}; ][]{kotonya2020explainable}. Extractive and abstractive summarization approaches still have many problems: extractive-generated explanations can not generate sufficiently context-full explanations, while abstractive-generated ones may lack faithfulness, given the tendency to hallucinate of these neural models \citep{kotonya2020survey, guo2022survey}. Currently, the abstractive summarization technique appears to be the most viable option for generating effective justifications. Nevertheless, it may not always be possible to acquire an adequate amount of training data or the necessary computational resources for highly demanding models. Thus, the purpose of this paper is twofold: (i) provide SOTA results using simple off-the-shelf LMs, and (ii) understand which is the most suitable approach for a given scenario.

\section{Datasets}
\label{datasets}

For our experiments, we collected two datasets with different structural and stylistic features. The first is LIAR++, a derivation of LIAR-PLUS \citep{alhindi-etal-2018-evidence}, while the second is FullFact, a completely new dataset.
Both datasets comprise claim, verdict, and article entries.

The \textit{claim} is a short text consisting of a statement that is under inspection: it can be \textsc{True}, partially \textsc{True},  or \textsc{False}. The \textit{verdict} is usually a paragraph-long text that provides arguments to assess the truth value of the claim: in many cases, it corresponds to a debunking text\footnote{In the literature, the term verdict often indicates the degree of truthfulness of a claim, and it is usually expressed as a label. Instead, our verdict contains also the so-called \textit{justification} or \textit{explanation} of the verdict label.}. Finally, the \textit{article} is a document that discusses the veracity of the claim using a journalistic style and contains the verifiable facts necessary to build the verdict. Figure \ref{fig:dialogue_example} illustrates an example for each element. A detailed description of the employed datasets follows\footnote{The code for dataset creation can be found at the following link \href{https://github.com/LanD-FBK/benchmark-gen-explanations}{https://github.com/LanD-FBK/benchmark-gen-explanations}.}.

\subsection{LIAR++ Dataset}

We created LIAR++ (L$_{++}$ henceforth) starting from the LIAR-PLUS dataset \cite{alhindi-etal-2018-evidence}. This dataset contains articles from the \textsc{PolitiFact} website\footnote{\href{https://www.politifact.com}{https://www.politifact.com}} spanning from 2007 to 2016 and covers various political topics with a primary emphasis on verifying the accuracy of statements made by political figures. LIAR-PLUS contains some entries in which the verdict was \textit{artificially created} by extracting the last five sentences from the body of the article. In all the other cases, verdicts were extracted from a specific section of web pages, usually titled \textit{Our ruling} or \textit{Summing up}. Qualitative and quantitative analyses of the \textit{artificial} against \textit{gold} verdicts showed that the former did not meet the expected quality. Therefore we decided to discard them while creating L$_{++}$. Differently from LIAR-PLUS, we also kept the whole verdict without removing the `forbidden sentences' (i.e., sentences comprising any verdict-related word) such as ``\textit{this statement is false}" \footnote{LIAR-PLUS was meant for claim classification, thereby those forbidden sentences would have made the task trivial.}. After this procedure L$_{++}$ comprises 6451 \textit{claim}-\textit{article}-\textit{verdict} triples.

\subsection{FullFact Dataset}

With a similar procedure to that used for L$_{++}$, we created a new dataset starting from the \textsc{FullFact} website\footnote{\href{https://fullfact.org}{https://fullfact.org}} (FF henceforth). This dataset contains data spanning from 2010 to 2021, and covers several different topics, such as health, economy, crime, law, and education. In FF the verdict is always present as a separate element in the web page so there was no need to filter the data. This dataset accounts for 1838 \textit{claim}-\textit{article}-\textit{verdict} triples.

\subsection{Analysis of the Datasets}

In this section, we focus on the main structural and stylistic differences between the two datasets, especially those that can have an impact on the experiments presented in the following sections. We mainly employed ROUGE score \citep{lin2004rouge} as evaluation metric in order to assess the quality of our datasets and of the generated summaries. ROUGE (Recall-Oriented Understudy for Gisting Evaluation) counts the number of overlapping units between two different texts. In the paper we reported: ROUGE-N (\textit{R-N}, N=1,2) which counts the number of n-grams overlapping, and ROUGE-L (\textit{R-L}) taking into account the LCS between two texts.

\paragraph{Average Article and Verdict Length.} Data length was computed in terms of the number of sentences, standard tokens, and BPE tokens\footnote{Computed using T5-large tokenizer}. As shown in Table \ref{table:lengths}, FF articles and verdicts are much shorter than the L$_{++}$ counterparts: 632 vs. 818 tokens and 30 vs. 114 tokens respectively. On the contrary, claim lengths are essentially similar (18 vs. 15). Regarding the lengths in terms of BPE tokens, the average length of articles alone exceeds the fixed input length of the major Language Models (LMs), which is usually 512 or 1024 (see Table \ref{table:lengths}). Indeed, 98\% and 54\% of L$_{++}$ articles are above the 512 and 1024 limit respectively, while 66\% and 24\% for FF. This implies that input reduction or truncation will be needed when processing the data during our experiments.


\begin{table}[h!]
\centering
\small
\begin{tabular}{llccc}
    \toprule
   \multicolumn{2}{c}{} & SENT$_{\mu}$ & TOK$_{\mu}$ & BPE$_{\mu}$ \\ 
  \midrule
  \multirow{3}{*}{L$_{++}$}   & Article & 38.9 & 817.8 & 1131.7 \\ 
                                                            & Claim & 1.2 & 17.9 & 24.9     \\
                                                            & Verdict & 6.3 & 113.7 & 150.4   \\
  \midrule

 \multirow{3}{*}{FF}    & Article & 24.8 & 632.1 & 803.5  \\ 
                                                      & Claim & 1.0 & 15.0 & 20.3  \\ 
                                                      & Verdict & 1.9 & 30.4 & 39.04  \\ 
  \bottomrule
\end{tabular}
\caption{Average length of each element of the datasets in terms of number of sentences (SENT), standard tokens (TOK) and BPE tokens (BPE).}
\label{table:lengths}
\end{table}


\paragraph{Presence of Verdict Snippets in the Article.} We compared the two datasets in terms of the possibility of abstracting/extracting the verdict from the article. In particular, we considered ROUGE recall to highlight how many verdict snippets are present in the article. Results indicate that L$_{++}$ has a more abstractive nature than FF  (see Table \ref{tab:verdict-article-overlap}). Indeed, the text of the verdict is present in the article more verbatim for FF than for L$_{++}$ (0.547 vs. 0.426 ROUGE-L recall). On the contrary, with ROUGE F1 we can observe how difficult it is to find verdict material in the article. Results show that FF articles contain very few pieces of FF verdicts. This can be explained in light of the much shorter length of the FF verdicts as compared to L$_{++}$ ones (39 vs. 150 BPE tokens on average, see Table \ref{table:lengths}), while article length difference is negligible in this comparison. 

To sum up, FF verdicts are much shorter than L$_{++}$ verdicts and even if they are present in longer verbatim sequences in the articles, these sequences are much more spread out the document. Thus, we expect that it will be harder to identify and extract FF verdicts.


\begin{table}[h!]
    \small
    \centering
    \small
    \begin{tabular}{llccc}
        \toprule
         &  & \multicolumn{1}{c}{R1} & \multicolumn{1}{c}{R2} & \multicolumn{1}{c}{RL}  \\ \midrule
        \multirow{2}{*}{L$_{++}$} &  Rec. 
        & 0.678 & 0.272 & 0.426 \\
        & F1 & 0.168 & 0.067 & 0.103\\
        \midrule
        \multirow{2}{*}{FF} & 
        Rec. & 0.724 & 0.355 & 0.547\\
        & F1 & 0.093 & 0.045 & 0.068\\
        \bottomrule
    \end{tabular}
    \caption{
    Verdict and article overlap measured in terms of ROUGE F1 and Recall scores.}
    \label{tab:verdict-article-overlap}
\end{table}


\paragraph{Claim Repetition in Verdict.} The possible presence of significant parts of the claim in the verdict positively affects the ROUGE scores without necessarily indicating a better verdict quality\footnote{e.g. ``\textit{The statement that [People vaccinated against Covid-19 may acquire immunodeficiency syndrome] was originally posted by ...}"}. For example, a trivial baseline that, given a claim, outputs a verdict that simply states ``\textit{It is not true that [claim]"} would obtain a high ROUGE score without producing any significant explanation to a verdict. Thus, we analysed claim and verdict overlap and reported the results in Table \ref{tab:claim-verdict-overlap}. Considering ROUGE-L, on average ~65\% of claim's subsequences are quoted verbatim in the verdict for the L$_{++}$ dataset, while only ~26\% for FF. The frequent reference to the claim at the beginning of the verdict can explain this outcome (Example in Appendix \ref{app:verdicts_style} in Table \ref{tab:ex_claim_verdict}). To check this hypothesis we re-computed ROUGE scores after removing the first sentence of the claim. We also repeated the test by removing the last sentence as a control condition. We observe that for L$_{++}$ ROUGE scores drop when evaluating the overlap between claim and verdict without the first sentence (i.e. ROUGE-1 goes from 0.709 to 0.394). On the other hand, this is not the case with the removal of the last sentence (R1 is 0.702), 
which corroborates our hypothesis.

To sum up, L$_{++}$ verdicts comprise a good amount of claim information, usually reported in the first sentence. However, this does not apply to FF. Additionally, L$_{++}$ verdicts end with a statement about claim veracity, usually in the form \textit{``We rate the claim \texttt{[TRUTH LABEL]}''}.


\begin{table}[b!]
\small
\centering
\small
\begin{tabular}{clccc}
\toprule
 & Verdict & R1   & R2   & RL \\ 
\midrule
\multirow{3}{*}{L$_{++}$} & comp. & 0.709 & 0.532  & 0.648 \\ 
& no 1st & 0.394 & 0.130 & 0.302  \\ 
& no last &  0.702 & 0.527 & 0.643 \\ 
\midrule
\multirow{3}{*}{FF} & comp. & 0.311 & 0.099 & 0.257 \\ 
& no 1st & 0.247 & 0.074 & 0.208 \\ 
& no last & 0.192 & 0.061 & 0.165 \\ 
\bottomrule
\end{tabular}
\caption{Recall of ROUGE scores between claims and verdicts. \textit{comp.} indicates scores compute on the whole verdict, while \textit{no 1st} and \textit{no last} indicates the removal of the first and last sentence respectively.}
\label{tab:claim-verdict-overlap}
\end{table}



 \begin{figure*}[t!]
 \centering
 {\includegraphics[width=0.8\textwidth]{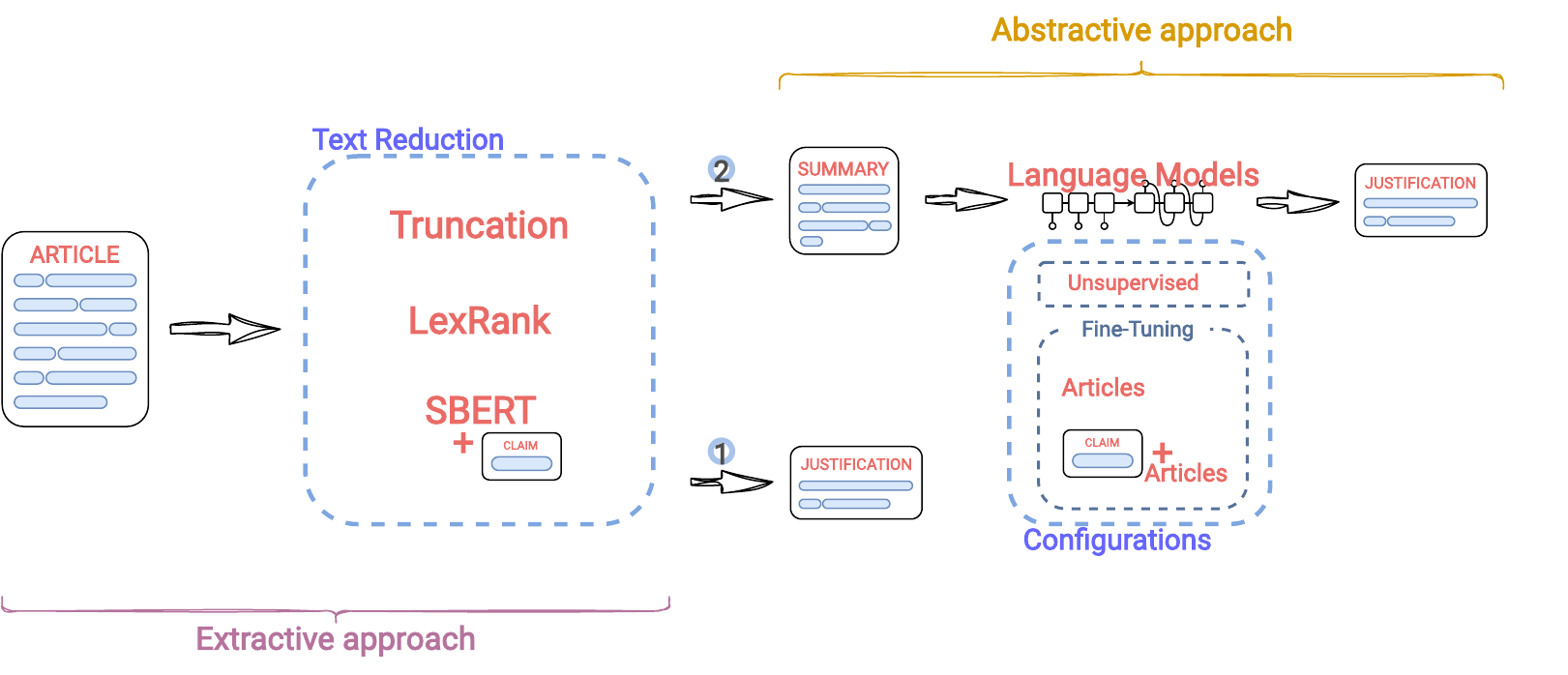}}
    \caption{General pipeline of our experiments for Justification Production. (1) extractive approach only; (2) extractive and abstractive summarization approaches combined in a unique pipeline. The total number of 960 configurations/experiments comprises 2 datasets (FF and L$_{++}$) $\times$ 10 summary configurations (8 from LexRank and SBERT both top/bottom and article/ranking order + 2 from Truncation head/tail) $\times$ 4 LMs (T5, dBart, Peg$_{xsum}$, Peg$_{cnn}$) $\times$ 3 fine-tuning (\texttt{unsupervised}, \texttt{article}, \texttt{claim+article}) $\times$ 4 decodings (beam search, Top-K sampling, nucleus sampling, and typical sampling).}
    \label{fig:pipeline_new}
 \end{figure*}


\paragraph{Article Adherence to the Claim.} The amount of text of the claim included in the article 
is a proxy for understanding: (i) if a simple summarization approach could provide a good verdict, even without explicitly providing the claim, and (ii) if there is a preferable portion of the article to be selected for summarization in order to fit into LMs' input length.

Since we know that the articles are written to discuss the veracity of the corresponding claims, we expect each article to contain 
a certain amount of information related to the claim, including partial or even whole quotations of it. This assumption would be reflected in high ROUGE recall values between the claim and the article. 

Results in Table \ref{tab:claim-article-overlap} confirm our expectations. While the high ROUGE-1 values can be trivially explained by the claim and article having the same topic, the high ROUGE-2 and L recall values indicate that entire portions of the claim were inserted into the article. On average, 80\% of the claim subsequences are quoted verbatim within the article in the two datasets. However, verbatim claim text is not particularly  
used in the first sentence of the article: its content is spread over the article, as can be seen by the small variation in ROUGE scores obtained by removing the first or last sentences of each document.

To sum up, the claim information is highly present within the article and spread over the entire text. For this reason, we expect extractive summarization approaches to be better than simple text truncation at selecting meaningful information from the article.


\begin{table}[t!]
\small
\centering
\small
\begin{tabular}{clccc}
\toprule
 & Article & R1   & R2   & RL  \\ 
\midrule

\multirow{3}{*}{L$_{++}$} & comp. & 0.875 & 0.706 & 0.810 \\ 
& no 1st & 0.862 & 0.689 & 0.791 \\ 
& no last & 0.874 & 0.704 & 0.808\\ 
\midrule
\multirow{3}{*}{FF} & comp. & 0.785 & 0.426 & 0.661\\ 
& no 1st & 0.739 & 0.351 & 0.612\\ 
& no last & 0.779 & 0.421 & 0.655\\ 
\bottomrule
\end{tabular}
\caption{Recall of ROUGE scores between claims and articles. \textit{comp.} indicates scores compute on the whole verdict, while \textit{no 1st} and \textit{no last} indicates the removal of the first and last sentence respectively.}
\label{tab:claim-article-overlap}
\end{table}


\section{Experimental Setup}

In this section, we present several experiments for the task of justification production. All the approaches can be traced back to the pipeline presented in Figure \ref{fig:pipeline_new}. Given an article, we tested several extractive approaches to select relevant material. Extractive summarises were considered Justifications per se or were sent to a Language Model (LM) pre-trained on the abstractive summarization objective. The LMs, in turn, were used with or without a fine-tuning step. Moreover, we selected different decoding mechanisms to drive the generation. Eventually, we conduct a cross-domain experiment to evaluate the models’ robustness to the style of each dataset.

\subsection{Extractive Approaches}

We first explored unsupervised extractive methods by comparing three different settings:
article truncation, article-relevance extractive summarization (using LexRank algorithm) and claim-driven extractive summarization (with SBERT). Each configuration represents a different assumption: (i) the main content (corresponding to a possible verdict) is introduced at the beginning or at the end of the article, within a specific section; (ii) a proper extractive summary or verdict contains the most relevant sentences of the article; (iii) a proper verdict comprises the article sentences most similar to the claim. 

\begin{itemize}[leftmargin=*]
    \itemsep0em
\item \textbf{Truncation} is the most straightforward approach of ``input reduction'', i.e. cutting the input at a given threshold. This is the simplest procedure applied when using LMs on long texts.

\item \textbf{LexRank} \cite{erkan2004lexrank} is an unsupervised approach for extractive text summarization which ranks the sentences of a document through a graph-based centrality scoring.

\item \textbf{SBERT} \citep{reimers2019sentence} is a siamese network based on BERT \citep{devlin2018bert} employed for generating and ranking sentence embeddings with respect to a target sentence (i.e. the claim) using cosine-similarity.
\end{itemize}

All the reduction baselines were tested under two configurations: from the list of sentences they provide, we selected either the top or bottom of the list. Furthermore, for LexRank and SBERT we rearranged top or bottom sentences according to article or ranking order.

\subsection{Abstractive Approach}
In the second part of our experimental design, we combined extractive and abstractive summarization for justification production. A reduced version of the text, obtained through truncation or extractive summarization, was used as input to various off-the-shelf Transformer-based models pre-trained on an abstractive summarization objective. In particular, we experiment with 4 Transformer-based summarization LMs\footnote{We have used Huggingface Transformers library for our experiments: \href{https://huggingface.co/transformers}{https://huggingface.co/transformers}.} trained on news-specific summarization datasets:

\begin{itemize}[leftmargin=*]
    \itemsep0em
    \item \textbf{T5}: T5-large, 738M parameters, input size 512, \citep{roberts2020exploring}
    \item \textbf{Peg$_{xsum}$}: Pegasus xsum, 570M parameters, input size 512 \citep{zhang2020pegasus}
    \item \textbf{Peg$_{cnn}$}: Pegasus cnn\_dailymail,  570M parameters, input size 1024 \citep{zhang2020pegasus}
    \item \textbf{dBart}: DistilBart cnn-12-6, 305M parameters, input size 1024 \citep{shleifer2020pre}
\end{itemize}

All the models were tested under three main configurations: \texttt{unsupervised}, fine-tuned on a reduced version of the article (\texttt{article}), and fine-tuned on the concatenation of the claim and the reduced article (\texttt{claim+article}). Finally, four decoding mechanisms were employed for generating the verdicts: \textit{beam search} (5 beams), \textit{Top-K sampling} (sampling pool limited to 40 words), \textit{nucleus sampling} (probability set to 0.9), and \textit{typical sampling} \citep[probability set to 0.95;][]{meister2022typical}. The fine-tuning details and hyperparameter settings can be found in Appendix \ref{App_fine-tuning}. 

\section{Experimental Results}

Our experimental design combines all the settings described in the previous sections. Extractive and abstractive approaches are concatenated in a unique pipeline tested on both L$_{++}$ and FF. Additionally, we tested the generalisation capabilities of the pipeline in zero-shot experiments and by integrating the two datasets into a unique model. Although we tested the complete design (960 configurations, as depicted in Figure \ref{fig:pipeline_new}), we will discuss only the most relevant findings hereafter. 

\paragraph{Claim-driven Extractive Summarization.} \label{subsec:extractive_summarization} If we focus on verdict generation as a pure unsupervised extractive summarization task, then \textit{the claim-driven approach through SBERT leads to better results in both datasets} (see Table \ref{tab:extractive_sum}). The second best approach is LexRank, which focuses on sentence relevance within the article (rather than claim relevance). Simple truncation led to the lowest results when considering ROUGE-1. In Table \ref{tab:extractive_sum} we report the best results, i.e. top selection with article order. Results for bottom selection and ranking order are reported in Table \ref{tab:extractive_sum_complete} in Appendix \ref{appendix_results} \footnote{Bottom approaches represent specific assumptions: (i) for truncation, the hypothesis is that informative content is in the last lines of the articles (in the form of a ‘‘to sum up’’ paragraph); (ii) for SBERT that the most similar sentences could be those that simply rephrase the claim but are not necessarily the most informative.}.


\begin{table}[h!]
    \centering
    \small
    \begin{tabular}{llccc}
        \toprule
        \multicolumn{1}{l}{} 
        & \multicolumn{1}{l}{Method} & R1    & R2    & RL    \\ \hline
            \multirow{3}{*}{FF}    
                & truncation & 0.258 & 0.082 & 0.182 \\
                & LexRank & 0.267 & 0.083 & 0.180  \\
                & SBERT & \textbf{0.300} & \textbf{0.114} & \textbf{0.213} \\ 
            \midrule
                \multirow{3}{*}{L$_{++}$}       
                & truncation & 0.347 & 0.120 & 0.196 \\
                & LexRank & 0.373 & 0.120 & 0.194  \\
                & SBERT & \textbf{0.393} & \textbf{0.158} & \textbf{0.237} \\ 
        \bottomrule
    \end{tabular}
    \caption{Extractive approaches comparison. The number of sentences to be extracted is set to the average number of sentences per verdict in the corresponding datasets (2 for FF and 6 for L$_{++}$).}
    \label{tab:extractive_sum}
\end{table}


\paragraph{Sentence Order for LM Input.} An aspect that can have a significant impact on LMs' performance is the order of the sentences fed to the LMs. Results show that \textit{rearranging sentences according to ranking order, rather than article order, can hinder text coherence}. As can be seen in Table \ref{tab:text_order_results}, article order is generally better than ranking order for 1024 input size LMs with L$_{++}$. For FF, article order leads to higher ROUGE scores with 512 input-size LMs. Differences among datasets can be explained by the lengths of their articles: in particular, most of the articles from FF are shorter than 512 BPE tokens.


\begin{table}[b!]
    \centering
    \small
    \begin{tabular}{lllccc}
        \toprule
                      & Model                       & Order   & R1    & R2    & RL    \\
        \midrule
        \multirow{8}{*}{\begin{sideways}L$_{++}$\end{sideways}}
                             & \multirow{2}{*}{T5}         & art. & 0.448 & 0.240 & 0.349 \\
                             &                             & rank. & 0.454 & 0.242 & 0.351 \\ \cmidrule(l){2-6}
                             & \multirow{2}{*}{Peg$_{xsum}$}       & art. & 0.452 & 0.247 & 0.355 \\
                             &                             & rank. & 0.455 & 0.249 & 0.357 \\ \cmidrule(l){2-6}
                             & \multirow{2}{*}{dBart}      & art.  & 0.460 & 0.254 & 0.359 \\
                             &                             & rank. & 0.454 & 0.246 & 0.352 \\ \cmidrule(l){2-6}
                             & \multirow{2}{*}{Peg$_{cnn}$}        & art. & \textbf{0.476} & \textbf{0.261} & \textbf{0.371} \\
                             &                             & rank. & 0.467 & 0.255 & 0.366 \\
        \midrule
        \multirow{8}{*}{\begin{sideways}FF\end{sideways}}
                             & \multirow{2}{*}{T5}         & art.  & 0.360 & 0.139 & 0.269 \\
                             &                             & rank. & 0.342 & 0.128 & 0.257 \\ \cmidrule(l){2-6}
                             & \multirow{2}{*}{Peg$_{xsum}$}       & art.  & \textbf{0.359} & \textbf{0.144} & \textbf{0.269} \\
                             &                             & rank. & 0.334 & 0.121 & 0.246 \\ \cmidrule(l){2-6}
                             & \multirow{2}{*}{dBart}      & art.  & 0.350 & 0.131 & 0.255 \\
                             &                             & rank. & 0.358 & 0.143 & 0.265 \\ \cmidrule(l){2-6}
                             & \multirow{2}{*}{Peg$_{cnn}$}        & art.  & 0.355 & 0.138 & 0.261 \\
                             &                             & rank. & 0.335 & 0.126 & 0.248 \\
        \bottomrule
    \end{tabular}
    \caption{ROUGE F1 scores for each model in the SBERT top \texttt{claim+article} configuration. Verdicts were generated through the beam search decoding method (the best among the 4 decoding mechanisms tested). The input length size for T5 and Peg$_{xsum}$ is 512, while for dBart and Peg$_{cnn}$ is 1024.}
    \label{tab:text_order_results}
\end{table}


\paragraph{Claim-driven Abstractive Summarization.} One major question when using LMs is whether the claim information is essential to drive the generation of the justification. Indeed, we should consider that (i) the sentences used as LM input are already selected according to the claim (SBERT) (ii) we are using gold articles (i.e. specifically written to debunk the given claim). \textit{Results show that the enrichment of the input with the claim information leads to ROUGE scores even higher than those obtained through a simple fine-tuning on the articles only} (see Table \ref{tab:ft_config_results}).


\begin{table}[t!]
\centering
\small
\begin{tabular}{lllccc}
\toprule
                      & \multicolumn{2}{l}{Configuration}       & R1    & R2    & RL    \\ \midrule
\multirow{12}{*}{\begin{sideways}L$_{++}$\end{sideways}} & \multirow{3}{*}{T5}    & unsup.        & 0.293 & 0.103 & 0.194 \\
                      &                             & art.          & 0.437 & 0.217 & 0.328 \\
                      &                             & cl+art.    & \textbf{0.448} & \textbf{0.240} & \textbf{0.349} \\ \cmidrule(l){2-6}
                      & \multirow{3}{*}{Peg$_{xsum}$}       & unsup.        & 0.206 & 0.056 & 0.138 \\
                      &                             & art.          & 0.442 & 0.230 & 0.339 \\
                      &                             & cl+art.    & \textbf{0.452} & \textbf{0.247} & \textbf{0.355} \\ \cmidrule(l){2-6}
                      & \multirow{3}{*}{dBart}      & unsup.        & 0.336 & 0.115 & 0.214 \\
                      &                             & art.          & 0.452 & 0.225 & 0.333 \\
                      &                             & cl+art.    & \textbf{0.460} & \textbf{0.254} & \textbf{0.359} \\ \cmidrule(l){2-6}
                      & \multirow{3}{*}{Peg$_{cnn}$}        & unsup.        & 0.267 & 0.097 & 0.189 \\
                      &                             & art.          & 0.463 & 0.238 & 0.350  \\
                      &                             & cl+art.    & \textbf{0.476} & \textbf{0.261} & \textbf{0.371} \\ \midrule
\multirow{12}{*}{\begin{sideways}FF\end{sideways}}  & \multirow{3}{*}{T5}         & unsup.        & 0.302 & 0.103 & 0.203 \\
                      &                             & art.          & 0.331 & 0.117 & 0.239  \\
                      &                             & cl+art.    & \textbf{0.360} & \textbf{0.139} & \textbf{0.269} \\ \cmidrule(l){2-6}
                      & \multirow{3}{*}{Peg$_{xsum}$}       & unsup.        & 0.241 & 0.061 & 0.174 \\
                      &                             & art.          & 0.329 & 0.125 & 0.244 \\
                      &                             & cl+art.    & \textbf{0.359} & \textbf{0.144} & \textbf{0.269} \\\cmidrule(l){2-6} 
                      & \multirow{3}{*}{dBart}       & unsup.       & 0.284 & 0.100 & 0.187 \\
                      &                             & art.          & 0.320 & 0.113 & 0.233  \\
                      &                             & cl+art.    & \textbf{0.350} & \textbf{0.131} & \textbf{0.255} \\ \cmidrule(l){2-6} 
                      & \multirow{3}{*}{Peg$_{cnn}$}        & unsup.        & 0.281 & 0.094 & 0.196 \\
                      &                             & art.          & 0.319 & 0.113 & 0.234  \\
                      &                             & cl+art.    & \textbf{0.355} & \textbf{0.138} & \textbf{0.261}  \\ \bottomrule
\end{tabular}
\caption{ROUGE F1 scores for each model in the SBERT top configuration. Results for both the unsupervised and the two fine-tuning settings (\texttt{article} and \texttt{claim+article}) are reported. Verdicts were generated through the beam search decoding method.}
\label{tab:ft_config_results}
\end{table}


\paragraph{LM Input Length.} Throughout the experiments, we saw that 1024 input-length models had higher results on L$_{++}$, while on FF better performances were recorded with 512 input-length models (see Table \ref{tab:input_length_analysis}). A possible explanation is that the differences in performance are due to the average length of articles in the two datasets (longer for L++, shorter for FF, see Table \ref{table:lengths}). In order to provide additional evidence for this hypothesis, we calculated ROUGE scores exclusively for articles with a length of 512 BPE tokens or less from both datasets. The results indicate that ROUGE scores for 512 input models were higher in both datasets than those obtained with 1024 \textit{proving that article length is the key factor when selecting the proper model}.


\begin{table}[t!]
\small
\centering
\begin{tabular}{@{}llcccc@{}}
\toprule
                   &                   & Length & R1    & R2    & RL    \\ \midrule
\multirow{6}{*}{L$_{++}$} & \multirow{2}{*}{unsup.}  & 512          & 0.249 & \textbf{0.121} & 0.166 \\
                          &                                & 1024         & \textbf{0.302} & 0.106 & \textbf{0.202} \\ \cmidrule(l){2-6}
                          & \multirow{2}{*}{art.}      & 512          & 0.440 & 0.224 & 0.334 \\
                          &                                & 1024         & \textbf{0.458} & \textbf{0.232} & \textbf{0.342} \\ \cmidrule(l){2-6}
                          & \multirow{2}{*}{cl+art.} & 512          & 0.450 & 0.244 & 0.352 \\
                          &                                & 1024         & \textbf{0.468} & \textbf{0.257} & \textbf{0.365} \\ \midrule
\multirow{6}{*}{FF}       & \multirow{2}{*}{unsup.}  & 512          & 0.272 & 0.082 & 0.189 \\
                          &                                & 1024         & \textbf{0.283} & \textbf{0.097} & \textbf{0.192} \\ \cmidrule(l){2-6}
                          & \multirow{2}{*}{art.}      & 512          & \textbf{0.330} & \textbf{0.121} & \textbf{0.242} \\
                          &                                & 1024         & 0.320 & 0.113 & 0.234 \\ \cmidrule(l){2-6}
                          & \multirow{2}{*}{cl+art.} & 512          & \textbf{0.360} & \textbf{0.142} & \textbf{0.269} \\
                          &                                & 1024         & 0.353 & 0.135 & 0.258 \\ \bottomrule 
\end{tabular}
\caption{Averaged results for models' input length. ROUGE scores for verdicts generated under the SBERT, article order, top, \texttt{claim+article} configuration (beam search decoding). }
\label{tab:input_length_analysis}
\end{table}



\begin{table}[b!]
\centering
\small
\begin{tabular}{p{0.95\linewidth}} 
 \toprule 
    \textbf{\texttt{claim}} : There have been 1,400 deaths and one million injured from Covid-19 vaccinations in the UK.\\
 \midrule
    \textbf{\texttt{gold verdict}} : These are deaths and potential side effects reported following the vaccine, not necessarily because of it. \\
 \midrule 
     \textbf{\texttt{SBERT}} : The front page of free newspaper `The Light', shared on Facebook, claimed that there have been 1,400 deaths and a million injuries ``from covid injections'' in the UK. There had been just over 1,470 deaths following a Covid-19 vaccination in the UK, according to the Medicines and Healthcare products Regulatory Agency's (MHRA) Yellow Card reporting scheme, as of 7 July 2021, when the paper came out.\\
 \midrule  
     \textbf{\texttt{abstractive}} : This is technically correct, but the fact that a death is reported following a vaccination is no proof the vaccine was the cause of this death or injury. \\
\bottomrule
\end{tabular}
\caption{FF example of generated verdicts. The first one is generated through extractive summarization only (with SBERT); the second example is the output of the extractive and abstractive pipeline (SBERT, top, article order, \texttt{claim+article} configuration with Peg$_{cnn}$).}
\label{table:FF_output}
\end{table}


\paragraph{Extractive vs Abstractive Summarization} In most cases, extractive summarization is better than unsupervised abstractive summarization (especially when claim-driven) in terms of ROUGE scores. Thus, \textit{if no training data is available, claim-driven extractive summarization is a viable solution}. On the other hand, \textit{when training data is available the best approach is to combine claim-driven abstractive and extractive summarization}.

\section{Cross-data and Mixed-data Experiments}

Next, we explored the impact of the datasets' stylistic characteristics in several training/test configurations. First, we conducted a zero-shot cross-dataset experiment, then we investigated the effect of combining the two datasets for training a single model.


\begin{table}[b!]
\centering
\small
\begin{tabular}{llccc}
\toprule
 &  & R1 & R2 & RL \\
 \midrule
\multirow{4}{*}{\shortstack[c]{L$_{++}$$\rightarrow$FF\\articles}} & T5            & 0.274	& 0.087	& 0.181 \\
                                                                    & Peg$_{xsum}$  & 0.277	& 0.100	& 0.195 \\
                                                                    & Peg$_{cnn}$  & 0.266	& 0.089	& 0.182 \\
                                                                    & dBart        & 0.266	& 0.093	& 0.174 \\
\cmidrule(l){2-5}
\multirow{4}{*}{\shortstack[c]{L$_{++}$$\rightarrow$FF\\claim+art}} & T5            & 0.288	& 0.092	& 0.194 \\
                                                                    & Peg$_{xsum}$  & 0.282	& 0.109	& 0.200 \\
                                                                    & Peg$_{cnn}$   & 0.286	& 0.105	& 0.198 \\
                                                                    & dBart         & 0.278	& 0.098	& 0.191 \\
\midrule
\multirow{4}{*}{\shortstack[c]{FF$\rightarrow$L$_{++}$\\articles}}  & T5            & 0.256	& 0.087	& 0.171 \\
                                                                    & Peg$_{xsum}$  & 0.248	& 0.084	& 0.164 \\
                                                                    & Peg$_{cnn}$   & 0.245	& 0.084	& 0.167 \\
                                                                    & dBart         & 0.269	& 0.081	& 0.172 \\
\cmidrule(l){2-5}
\multirow{4}{*}{\shortstack[c]{FF$\rightarrow$L$_{++}$\\claim+art.}}    & T5            & 0.271	& 0.099	& 0.184 \\
                                                                        & Peg$_{xsum}$  & 0.262	& 0.099	& 0.178 \\
                                                                        & Peg$_{cnn}$   & 0.244	& 0.087	& 0.168 \\
                                                                        & dBart         & 0.274	& 0.092	& 0.180 \\
\bottomrule
\end{tabular}
\caption{Models fine-tuned on FF and tested on Liar+ test set (FF$\rightarrow$L$_{++}$) and fine-tuned on L$_{++}$ and tested on FF test set (L$_{++}$$\rightarrow$FF), using \texttt{article} or \texttt{claim+article} configurations (SBERT top article order).}
\label{tab:cross_dataset_exp}
\end{table}


\paragraph{Cross-dataset Experiments.} In these experiments, the models were fine-tuned on one dataset and tested on the other, i.e. fine-tuned on L$_{++}$ and tested on FF (L$_{++}$$\rightarrow$FF) and vice-versa (FF$\rightarrow$L$_{++}$) both for \texttt{article} and \texttt{claim+article} configurations. In particular, we employed the best-performing pipeline from the previous experiments, i.e. models from the SBERT, top, article order configuration. Results are reported in Table \ref{tab:cross_dataset_exp}. Both for L$_{++}$ and FF, the models show the trend highlighted previously: the \texttt{claim+article} configuration performs better than the \texttt{article} configuration. Furthermore, as expected, testing on a different dataset yielded lower results: in several cases, results for the \texttt{article} configuration were on par or even worse than those obtained with the unsupervised LMs (compare with Table \ref{tab:ft_config_results}). This is particularly evident for the FF$\rightarrow$L$_{++}$ configuration. The low ROUGE values can be attributed to the distinct styles of the datasets and not to any degradation in the generation quality. As can be seen from the examples in Table \ref{tab:cross_examples}, models fine-tuned on L$_{++}$, even when tested on FF, generate justifications mimicking L$_{++}$ style (claim in the first sentence and truthfulness statement at the end, see Appendix \ref{app:verdicts_style}), and vice versa (FF$\rightarrow$L$_{++}$).


\begin{table}[h!]
\centering
\small
\begin{tabular}{p{0.93\linewidth}}
\toprule
    \multicolumn{1}{c}{L$_{++}$$\rightarrow$FF} \\ \midrule
    \textbf{\texttt{claim}}: University applications have fallen by 4\%. \\
    \textbf{\texttt{gold verdict}}: The number of applicants through UCAS to UK higher education institutions has gone down by 4\% compared to last year. \\
    \textbf{\texttt{generated verdict}}: The total number of people applying to higher education in the UK this year (as of the final June 2017 deadline) is down 4\% compared to last year, driven by falls from England, Wales and the EU. There have been a number of reasons for the overall drop, including a fall in the number of 18 year olds in the UK population. We rate this claim True. \\ 
    \midrule
    \multicolumn{1}{c}{FF$\rightarrow$L$_{++}$} \\ \midrule
    \textbf{\texttt{claim}}: Muslim nations did not call out the people who celebrated the 9/11 attacks. \\
    \textbf{\texttt{gold verdict}}: O'Reilly said that Muslim nations did not ``call out" the people who celebrated the 9/11 attacks. So far as we can tell, there was no official condemnation of people celebrating the 9/11 attacks. However, Muslim governments, and religious leaders, condemned the attacks themselves, as did many average Muslims. O'Reilly's statement is accurate as far as it goes, but it leaves out important information that might lead someone to reach a different conclusion. That meets our definition of Half True. \\
    \textbf{\texttt{generated verdict}}: There was no official condemnation from a Muslim-majority nation. What we did find were many official condemnations of the attacks themselves. Average Muslims attended candlelight vigils and other public events to voice sympathy for the victims and to repudiate the attacks. \\ \bottomrule
\end{tabular} 
\caption{Examples from the cross-data experiments.}
\label{tab:cross_examples}
\end{table}

\paragraph{Mixed Data Experiments.} 

Finally, we tested the effect of using both datasets in training a single LM. We focused on Peg$_{cnn}$ as in the in-domain experiments it generally showed quantitatively (see Table \ref{tab:ft_config_results}) and qualitatively (see Table \ref{table:FF_output}) better results, especially for longer input sequences. To this end, we combined the training data in a unique unbalanced dataset and we tested the extractive and abstractive pipeline (SBERT, top, article order setting). Results,  reported in Table \ref{tab:unbalanced-exp}, were found to be comparable to those achieved through the in-domain fine-tuning of distinct models for each dataset (see Table \ref{tab:ft_config_results}). Thus, if the datasets have peculiar styles, a more efficient way to tackle the task is to fine-tune a unique LM on all the data available rather than fine-tuning different models for each dataset.



\begin{table}[h!]
\centering
\small
\begin{tabular}{lccc}
\toprule
 & R1 & R2 & RL \\
\midrule
    L$_{++}$    & 0.473	& 0.261	& 0.370 \\
    FF          & 0.367	& 0.143	& 0.272 \\
\bottomrule
\end{tabular}
\caption{F1 ROUGE scores of Peg$_{cnn}$ fine-tuned on a unique dataset and tested on L$_{++}$ and FF test sets.}
\label{tab:unbalanced-exp}
\end{table}


\section{Conclusions}

Curbing misinformation with NLP tools is a crucial task. Up to now, researchers have mainly focused on claim veracity classification. In this paper, instead, we focused on generating textual justifications with factual and objective information to support a verdict. We started casting the problem as a news article summarization task and subsequently we integrated summarization within an end-to-end claim-driven explanation generation framework, accounting for the several practical scenarios that can be encountered. To this end, we experimented with several extractive and abstractive approaches, leveraging pre-trained LMs under manifold configurations. In order to provide an exhaustive benchmark of the justification production task, we employed two novel datasets throughout the experiments. The main results show that summarization needs to be driven by the claim to obtain better performances and that an extractive step before LM abstractive summarization further improves the results. Finally, we show that style information can be retained by a single model which is able to handle multiple datasets at once.

\section*{Limitations}

LMs suffer from hallucination \citep{zellers2019defending, solaiman2019release} and, even if the phenomenon is reduced by the document-driven nature of the task, it is still present. In particular, some hallucinations are critical: we occasionally obtain the sentence "\textit{we rate this statement as false}" even if the statement is true since it is a very common sentence in the L${_{++}}$ training set. 

Moreover, the datasets used for this task (i) are restricted to the English language and (ii) assume that there is always a gold article for fact-checking. In real scenarios we might have the debunking material spread over several articles: in this case, we can expect that models not suffering from the input size limit would be most beneficial. Still, from preliminary experiments, we conducted with two long input LMs on our datasets, namely LED-Large \citep{beltagy2020longformer} and \textsc{BertSumExtAbs} \citep{liu2019text} from \citet{kotonya2020explainable}, results were worse also for articles exceeding the 1024 input limit.

Another aspect that should be addressed is an in-depth analysis of automatically generated verdicts and their persuasiveness. In fact, different versions of a verdict for the same claim can have different effects depending on the audience -- e.g. for some people explanations comprising few arguments are more effective than longer explanations  \cite{sanna2006metacognitive}. To this end, carefully designed human evaluation experiments are needed.

\section*{Acknowledgements}
We would like to thank our TACL Action Editor and the three anonymous reviewers for their constructive feedback during the review process. This work was partly supported by the AI4TRUST project - AI-based-technologies for trustworthy solutions against disinformation (ID: 101070190).
\bibliography{tacl2021,custom}

\begin{thebibliography}{44}
\expandafter\ifx\csname natexlab\endcsname\relax\def\natexlab#1{#1}\fi

\bibitem[{Ahmadi et~al.(2019)Ahmadi, Lee, Papotti, and
  Saeed}]{ahmadi2019explainable}
Naser Ahmadi, Joohyung Lee, Paolo Papotti, and Mohammed Saeed. 2019.
\newblock \href {https://doi.org/https://doi.org/10.48550/arXiv.1906.09198}
  {Explainable fact checking with probabilistic answer set programming}.
\newblock \emph{ArXiv}, cs.DB/1906.09198.
\newblock Version 1.

\bibitem[{Alhindi et~al.(2018)Alhindi, Petridis, and
  Muresan}]{alhindi-etal-2018-evidence}
Tariq Alhindi, Savvas Petridis, and Smaranda Muresan. 2018.
\newblock \href {https://doi.org/10.18653/v1/W18-5513} {Where is your evidence:
  Improving fact-checking by justification modeling}.
\newblock In \emph{Proceedings of the First Workshop on Fact Extraction and
  {VER}ification ({FEVER})}, pages 85--90, Brussels, Belgium. Association for
  Computational Linguistics.

\bibitem[{Atanasova et~al.(2020)Atanasova, Simonsen, Lioma, and
  Augenstein}]{atanasova2020generating}
Pepa Atanasova, Jakob~Grue Simonsen, Christina Lioma, and Isabelle Augenstein.
  2020.
\newblock \href {https://doi.org/10.18653/v1/2020.acl-main.656} {Generating
  fact checking explanations}.
\newblock In \emph{Proceedings of the 58th Annual Meeting of the Association
  for Computational Linguistics}, pages 7352--7364, Online. Association for
  Computational Linguistics.

\bibitem[{Augenstein et~al.(2019)Augenstein, Lioma, Wang, Chaves~Lima, Hansen,
  Hansen, and Simonsen}]{augenstein2019multifc}
Isabelle Augenstein, Christina Lioma, Dongsheng Wang, Lucas Chaves~Lima, Casper
  Hansen, Christian Hansen, and Jakob~Grue Simonsen. 2019.
\newblock \href {https://doi.org/10.18653/v1/D19-1475} {{M}ulti{FC}: A
  real-world multi-domain dataset for evidence-based fact checking of claims}.
\newblock In \emph{Proceedings of the 2019 Conference on Empirical Methods in
  Natural Language Processing and the 9th International Joint Conference on
  Natural Language Processing (EMNLP-IJCNLP)}, pages 4685--4697, Hong Kong,
  China. Association for Computational Linguistics.

\bibitem[{Beltagy et~al.(2020)Beltagy, Peters, and
  Cohan}]{beltagy2020longformer}
Iz~Beltagy, Matthew~E. Peters, and Arman Cohan. 2020.
\newblock \href {https://doi.org/https://doi.org/10.48550/arXiv.2004.05150}
  {Longformer: The long-document transformer}.
\newblock \emph{arXiv}, cs.CL/2004.05150.
\newblock Version 2.

\bibitem[{Chen and Sharma(2015)}]{chen2015learning}
Rui Chen and Sushil~K. Sharma. 2015.
\newblock \href {https://doi.org/https://doi.org/10.1057/ejis.2013.31}
  {Learning and self-disclosure behavior on social networking sites: the case
  of facebook users}.
\newblock \emph{European Journal of Information Systems}, 24:93--106.

\bibitem[{Devlin et~al.(2019)Devlin, Chang, Lee, and
  Toutanova}]{devlin2018bert}
Jacob Devlin, Ming-Wei Chang, Kenton Lee, and Kristina Toutanova. 2019.
\newblock \href {https://doi.org/10.18653/v1/N19-1423} {{BERT}: Pre-training of
  deep bidirectional transformers for language understanding}.
\newblock In \emph{Proceedings of the 2019 Conference of the North {A}merican
  Chapter of the Association for Computational Linguistics: Human Language
  Technologies, Volume 1 (Long and Short Papers)}, pages 4171--4186,
  Minneapolis, Minnesota. Association for Computational Linguistics.

\bibitem[{Erkan and Radev(2004)}]{erkan2004lexrank}
G\"{u}nes Erkan and Dragomir~R. Radev. 2004.
\newblock \href {https://dl.acm.org/doi/10.5555/1622487.1622501} {Lexrank:
  Graph-based lexical centrality as salience in text summarization}.
\newblock \emph{J. Artif. Int. Res.}, 22(1):457–479.

\bibitem[{Gad-Elrab et~al.(2019)Gad-Elrab, Stepanova, Urbani, and
  Weikum}]{gad2019exfakt}
Mohamed~H. Gad-Elrab, Daria Stepanova, Jacopo Urbani, and Gerhard Weikum. 2019.
\newblock \href {https://doi.org/10.1145/3289600.3290996} {Exfakt: A framework
  for explaining facts over knowledge graphs and text}.
\newblock In \emph{Proceedings of the Twelfth ACM International Conference on
  Web Search and Data Mining}, WSDM '19, page 87–95, New York, NY, USA.
  Association for Computing Machinery.

\bibitem[{Glenski et~al.(2017)Glenski, Pennycuff, and
  Weninger}]{glenski2017consumers}
Maria Glenski, Corey Pennycuff, and Tim Weninger. 2017.
\newblock \href {https://doi.org/10.1109/TCSS.2017.2742242} {Consumers and
  curators: Browsing and voting patterns on reddit}.
\newblock \emph{IEEE Transactions on Computational Social Systems},
  4(4):196--206.

\bibitem[{Glenski et~al.(2020)Glenski, Volkova, and Kumar}]{glenski2020user}
Maria Glenski, Svitlana Volkova, and Srijan Kumar. 2020.
\newblock \href {https://doi.org/10.1007/978-3-030-42699-6_3} {\emph{User
  Engagement with Digital Deception}}. Springer International Publishing, Cham.

\bibitem[{Guo et~al.(2022)Guo, Schlichtkrull, and Vlachos}]{guo2022survey}
Zhijiang Guo, Michael Schlichtkrull, and Andreas Vlachos. 2022.
\newblock \href {https://doi.org/10.1162/tacl_a_00454} {A survey on automated
  fact-checking}.
\newblock \emph{Transactions of the Association for Computational Linguistics},
  10:178--206.

\bibitem[{Hassan et~al.(2015)Hassan, Adair, Hamilton, Li, Tremayne, Yang, and
  Yu}]{hassan2015quest}
Naeemul Hassan, Bill Adair, James~T. Hamilton, Chengkai Li, Mark Tremayne, Jun
  Yang, and Cong Yu. 2015.
\newblock The quest to automate fact-checking.
\newblock In \emph{Proceedings of the 2015 computation+ journalism symposium}.
  Citeseer.

\bibitem[{He et~al.(2023)He, Ahamad, and Kumar}]{he2023reinforcement}
Bing He, Mustaque Ahamad, and Srijan Kumar. 2023.
\newblock \href {https://doi.org/10.1145/3543507.3583388} {Reinforcement
  learning-based counter-misinformation response generation: A case study of
  covid-19 vaccine misinformation}.
\newblock In \emph{Proceedings of the ACM Web Conference 2023}, WWW '23, page
  2698–2709, New York, NY, USA. Association for Computing Machinery.

\bibitem[{Kotonya and Toni(2020{\natexlab{a}})}]{kotonya2020survey}
Neema Kotonya and Francesca Toni. 2020{\natexlab{a}}.
\newblock \href {https://doi.org/10.18653/v1/2020.coling-main.474} {Explainable
  automated fact-checking: A survey}.
\newblock In \emph{Proceedings of the 28th International Conference on
  Computational Linguistics}, pages 5430--5443, Barcelona, Spain (Online).
  International Committee on Computational Linguistics.

\bibitem[{Kotonya and Toni(2020{\natexlab{b}})}]{kotonya2020explainable}
Neema Kotonya and Francesca Toni. 2020{\natexlab{b}}.
\newblock \href {https://doi.org/10.18653/v1/2020.emnlp-main.623} {Explainable
  automated fact-checking for public health claims}.
\newblock In \emph{Proceedings of the 2020 Conference on Empirical Methods in
  Natural Language Processing (EMNLP)}, pages 7740--7754, Online. Association
  for Computational Linguistics.

\bibitem[{Lewandowsky et~al.(2012)Lewandowsky, Ecker, Seifert, Schwarz, and
  Cook}]{lewandowsky2012misinformation}
Stephan Lewandowsky, Ullrich K.~H. Ecker, Colleen~M. Seifert, Norbert Schwarz,
  and John Cook. 2012.
\newblock \href {https://doi.org/10.1177/1529100612451018} {Misinformation and
  its correction: Continued influence and successful debiasing}.
\newblock \emph{Psychological Science in the Public Interest}, 13(3):106--131.
\newblock PMID: 26173286.

\bibitem[{Lin(2004)}]{lin2004rouge}
Chin-Yew Lin. 2004.
\newblock \href {https://aclanthology.org/W04-1013} {{ROUGE}: A package for
  automatic evaluation of summaries}.
\newblock In \emph{Text Summarization Branches Out}, pages 74--81, Barcelona,
  Spain. Association for Computational Linguistics.

\bibitem[{Van~der Linden(2022)}]{van2022misinformation}
Sander Van~der Linden. 2022.
\newblock \href {https://doi.org/10.1038/s41591-022-01713-6} {Misinformation:
  Susceptibility, spread, and interventions to immunize the public}.
\newblock \emph{Nature Medicine}, 28(3):460--467.

\bibitem[{Liu and Lapata(2019)}]{liu2019text}
Yang Liu and Mirella Lapata. 2019.
\newblock \href {https://doi.org/10.18653/v1/D19-1387} {Text summarization with
  pretrained encoders}.
\newblock In \emph{Proceedings of the 2019 Conference on Empirical Methods in
  Natural Language Processing and the 9th International Joint Conference on
  Natural Language Processing (EMNLP-IJCNLP)}, pages 3730--3740, Hong Kong,
  China. Association for Computational Linguistics.

\bibitem[{Lombrozo(2007)}]{lombrozo2007simplicity}
Tania Lombrozo. 2007.
\newblock \href {https://doi.org/10.1016/j.cogpsych.2006.09.006} {Simplicity
  and probability in causal explanation}.
\newblock \emph{Cognitive psychology}, 55(3):232--257.

\bibitem[{Lu and Li(2020)}]{lu2020gcan}
Yi-Ju Lu and Cheng-Te Li. 2020.
\newblock \href {https://doi.org/10.18653/v1/2020.acl-main.48} {{GCAN}:
  Graph-aware co-attention networks for explainable fake news detection on
  social media}.
\newblock In \emph{Proceedings of the 58th Annual Meeting of the Association
  for Computational Linguistics}, pages 505--514, Online. Association for
  Computational Linguistics.

\bibitem[{Meister et~al.(2023)Meister, Pimentel, Wiher, and
  Cotterell}]{meister2022typical}
Clara Meister, Tiago Pimentel, Gian Wiher, and Ryan Cotterell. 2023.
\newblock \href {https://doi.org/10.1162/tacl_a_00536} {{Locally Typical
  Sampling}}.
\newblock \emph{Transactions of the Association for Computational Linguistics},
  11:102--121.

\bibitem[{Micallef et~al.(2020)Micallef, He, Kumar, Ahamad, and
  Memon}]{Micallef_et_al}
Nicholas Micallef, Bing He, Srijan Kumar, Mustaque Ahamad, and Nasir~D. Memon.
  2020.
\newblock \href {http://arxiv.org/abs/2011.05773} {The role of the crowd in
  countering misinformation: {A} case study of the {COVID-19} infodemic}.
\newblock \emph{arXiv}, cs.SI/2011.05773.
\newblock Version 2.

\bibitem[{Nakashole and Mitchell(2014)}]{nakashole2014language}
Ndapandula Nakashole and Tom~M. Mitchell. 2014.
\newblock \href {https://doi.org/10.3115/v1/P14-1095} {Language-aware truth
  assessment of fact candidates}.
\newblock In \emph{Proceedings of the 52nd Annual Meeting of the Association
  for Computational Linguistics (Volume 1: Long Papers)}, pages 1009--1019,
  Baltimore, Maryland. Association for Computational Linguistics.

\bibitem[{Ostrowski et~al.(2021)Ostrowski, Arora, Atanasova, and
  Augenstein}]{ijcai2021p536}
Wojciech Ostrowski, Arnav Arora, Pepa Atanasova, and Isabelle Augenstein. 2021.
\newblock \href {https://doi.org/10.24963/ijcai.2021/536} {Multi-hop fact
  checking of political claims}.
\newblock In \emph{Proceedings of the Thirtieth International Joint Conference
  on Artificial Intelligence, {IJCAI-21}}, pages 3892--3898. International
  Joint Conferences on Artificial Intelligence Organization.
\newblock Main Track.

\bibitem[{Pernice et~al.(2019)Pernice, Whitenton, and
  Nielsen}]{pernice2018people}
Kara Pernice, Kathryn Whitenton, and Jakob Nielsen. 2019.
\newblock \href
  {https://www.nngroup.com/reports/how-people-read-web-eyetracking-evidence/}
  {\emph{How People Read Online: The Eyetracking Evidence}}, 2nd edition.
\newblock Nielsen Norman Group.

\bibitem[{Popat et~al.(2018)Popat, Mukherjee, Yates, and
  Weikum}]{popat2018declare}
Kashyap Popat, Subhabrata Mukherjee, Andrew Yates, and Gerhard Weikum. 2018.
\newblock \href {https://doi.org/10.18653/v1/D18-1003} {{D}e{C}lar{E}:
  Debunking fake news and false claims using evidence-aware deep learning}.
\newblock In \emph{Proceedings of the 2018 Conference on Empirical Methods in
  Natural Language Processing}, pages 22--32, Brussels, Belgium. Association
  for Computational Linguistics.

\bibitem[{Potthast et~al.(2018)Potthast, Kiesel, Reinartz, Bevendorff, and
  Stein}]{potthast2017stylometric}
Martin Potthast, Johannes Kiesel, Kevin Reinartz, Janek Bevendorff, and Benno
  Stein. 2018.
\newblock \href {https://doi.org/10.18653/v1/P18-1022} {A stylometric inquiry
  into hyperpartisan and fake news}.
\newblock In \emph{Proceedings of the 56th Annual Meeting of the Association
  for Computational Linguistics (Volume 1: Long Papers)}, pages 231--240,
  Melbourne, Australia. Association for Computational Linguistics.

\bibitem[{Raffel et~al.(2020)Raffel, Shazeer, Roberts, Lee, Narang, Matena,
  Zhou, Li, and Liu}]{roberts2020exploring}
Colin Raffel, Noam Shazeer, Adam Roberts, Katherine Lee, Sharan Narang, Michael
  Matena, Yanqi Zhou, Wei Li, and Peter~J. Liu. 2020.
\newblock \href {http://jmlr.org/papers/v21/20-074.html} {Exploring the limits
  of transfer learning with a unified text-to-text transformer}.
\newblock \emph{Journal of Machine Learning Research}, 21(140):1--67.

\bibitem[{Reimers and Gurevych(2019)}]{reimers2019sentence}
Nils Reimers and Iryna Gurevych. 2019.
\newblock \href {https://doi.org/10.18653/v1/D19-1410} {Sentence-{BERT}:
  Sentence embeddings using {S}iamese {BERT}-networks}.
\newblock In \emph{Proceedings of the 2019 Conference on Empirical Methods in
  Natural Language Processing and the 9th International Joint Conference on
  Natural Language Processing (EMNLP-IJCNLP)}, pages 3982--3992, Hong Kong,
  China. Association for Computational Linguistics.

\bibitem[{Saakyan et~al.(2021)Saakyan, Chakrabarty, and
  Muresan}]{saakyan-etal-2021-covid}
Arkadiy Saakyan, Tuhin Chakrabarty, and Smaranda Muresan. 2021.
\newblock \href {https://doi.org/10.18653/v1/2021.acl-long.165} {{COVID}-fact:
  Fact extraction and verification of real-world claims on {COVID}-19
  pandemic}.
\newblock In \emph{Proceedings of the 59th Annual Meeting of the Association
  for Computational Linguistics and the 11th International Joint Conference on
  Natural Language Processing (Volume 1: Long Papers)}, pages 2116--2129,
  Online. Association for Computational Linguistics.

\bibitem[{Sanna and Schwarz(2006)}]{sanna2006metacognitive}
Lawrence~J. Sanna and Norbert Schwarz. 2006.
\newblock \href {https://doi.org/10.1111/j.1467-8721.2006.00430.x}
  {Metacognitive experiences and human judgment: The case of hindsight bias and
  its debiasing}.
\newblock \emph{Current Directions in Psychological Science}, 15(4):172--176.

\bibitem[{Shazeer and Stern(2018)}]{shazeer2018adafactor}
Noam Shazeer and Mitchell Stern. 2018.
\newblock \href {https://proceedings.mlr.press/v80/shazeer18a.html} {Adafactor:
  Adaptive learning rates with sublinear memory cost}.
\newblock In \emph{Proceedings of the 35th International Conference on Machine
  Learning}, volume~80 of \emph{Proceedings of Machine Learning Research},
  pages 4596--4604. PMLR.

\bibitem[{Shleifer and Rush(2020)}]{shleifer2020pre}
Sam Shleifer and Alexander~M. Rush. 2020.
\newblock \href {https://doi.org/https://doi.org/10.48550/arXiv.2010.13002}
  {Pre-trained summarization distillation}.
\newblock \emph{arXiv}, cs.CL/2010.13002.
\newblock Version 2.

\bibitem[{Shu et~al.(2019)Shu, Cui, Wang, Lee, and Liu}]{shu2019defend}
Kai Shu, Limeng Cui, Suhang Wang, Dongwon Lee, and Huan Liu. 2019.
\newblock \href {https://doi.org/10.1145/3292500.3330935} {Defend: Explainable
  fake news detection}.
\newblock In \emph{Proceedings of the 25th ACM SIGKDD International Conference
  on Knowledge Discovery \&amp; Data Mining}, KDD '19, page 395–405, New
  York, NY, USA. Association for Computing Machinery.

\bibitem[{Solaiman et~al.(2019)Solaiman, Brundage, Clark, Askell, Herbert-Voss,
  Wu, Radford, Krueger, Kim, Kreps, McCain, Newhouse, Blazakis, McGuffie, and
  Wang}]{solaiman2019release}
Irene Solaiman, Miles Brundage, Jack Clark, Amanda Askell, Ariel Herbert-Voss,
  Jeff Wu, Alec Radford, Gretchen Krueger, Jong~Wook Kim, Sarah Kreps, Miles
  McCain, Alex Newhouse, Jason Blazakis, Kris McGuffie, and Jasmine Wang. 2019.
\newblock \href {https://doi.org/https://doi.org/10.48550/arXiv.1908.09203}
  {Release strategies and the social impacts of language models}.
\newblock Version 2.

\bibitem[{Thorne et~al.(2018)Thorne, Vlachos, Christodoulopoulos, and
  Mittal}]{thorne2018fever}
James Thorne, Andreas Vlachos, Christos Christodoulopoulos, and Arpit Mittal.
  2018.
\newblock \href {https://doi.org/10.18653/v1/N18-1074} {{FEVER}: A large-scale
  dataset for fact extraction and {VER}ification}.
\newblock In \emph{Proceedings of the 2018 Conference of the North {A}merican
  Chapter of the Association for Computational Linguistics: Human Language
  Technologies, Volume 1 (Long Papers)}, pages 809--819, New Orleans,
  Louisiana. Association for Computational Linguistics.

\bibitem[{Wadden et~al.(2020)Wadden, Lin, Lo, Wang, van Zuylen, Cohan, and
  Hajishirzi}]{wadden-etal-2020-fact}
David Wadden, Shanchuan Lin, Kyle Lo, Lucy~Lu Wang, Madeleine van Zuylen, Arman
  Cohan, and Hannaneh Hajishirzi. 2020.
\newblock \href {https://doi.org/10.18653/v1/2020.emnlp-main.609} {Fact or
  fiction: Verifying scientific claims}.
\newblock In \emph{Proceedings of the 2020 Conference on Empirical Methods in
  Natural Language Processing (EMNLP)}, pages 7534--7550, Online. Association
  for Computational Linguistics.

\bibitem[{Wang(2017)}]{wang-2017-liar}
William~Yang Wang. 2017.
\newblock \href {https://doi.org/10.18653/v1/P17-2067} {{``L}iar, liar pants on
  fire{''}: {A} new benchmark dataset for fake news detection}.
\newblock In \emph{Proceedings of the 55th Annual Meeting of the Association
  for Computational Linguistics (Volume 2: Short Papers)}, pages 422--426,
  Vancouver, Canada. Association for Computational Linguistics.

\bibitem[{Wintersieck(2017)}]{Wintersieck}
Amanda~L. Wintersieck. 2017.
\newblock \href {https://doi.org/10.1177/1532673X16686555} {Debating the truth:
  The impact of fact-checking during electoral debates}.
\newblock \emph{American Politics Research}, 45(2):304--331.

\bibitem[{Yang et~al.(2019)Yang, Pentyala, Mohseni, Du, Yuan, Linder, Ragan,
  Ji, and Hu}]{yang2019xfake}
Fan Yang, Shiva~K. Pentyala, Sina Mohseni, Mengnan Du, Hao Yuan, Rhema Linder,
  Eric~D. Ragan, Shuiwang Ji, and Xia~(Ben) Hu. 2019.
\newblock \href {https://doi.org/10.1145/3308558.3314119} {Xfake: Explainable
  fake news detector with visualizations}.
\newblock In \emph{The World Wide Web Conference}, WWW '19, page 3600–3604,
  New York, NY, USA. Association for Computing Machinery.

\bibitem[{Zellers et~al.(2019)Zellers, Holtzman, Rashkin, Bisk, Farhadi,
  Roesner, and Choi}]{zellers2019defending}
Rowan Zellers, Ari Holtzman, Hannah Rashkin, Yonatan Bisk, Ali Farhadi,
  Franziska Roesner, and Yejin Choi. 2019.
\newblock \href
  {http://papers.nips.cc/paper/9106-defending-against-neural-fake-news.pdf}
  {Defending against neural fake news}.
\newblock In H.~Wallach, H.~Larochelle, A.~Beygelzimer, F.~d\textquotesingle
  Alch\'{e}-Buc, E.~Fox, and R.~Garnett, editors, \emph{Advances in Neural
  Information Processing Systems 32}, pages 9054--9065. Curran Associates, Inc.

\bibitem[{Zhang et~al.(2020)Zhang, Zhao, Saleh, and Liu}]{zhang2020pegasus}
Jingqing Zhang, Yao Zhao, Mohammad Saleh, and Peter Liu. 2020.
\newblock \href {https://proceedings.mlr.press/v119/zhang20ae.html} {{PEGASUS}:
  Pre-training with extracted gap-sentences for abstractive summarization}.
\newblock In \emph{Proceedings of the 37th International Conference on Machine
  Learning}, volume 119 of \emph{Proceedings of Machine Learning Research},
  pages 11328--11339. PMLR.

\end{thebibliography}
\bibliographystyle{acl_natbib}

\newpage
\appendix

\section{Verdict Stylistic Features}
\label{app:verdicts_style} 

Differently from FF, L$_{++}$ verdicts show a peculiar and recurrent style: the first sentence comprises a reference to the claim, usually quoted verbatim (see Table \ref{tab:ex_claim_verdict}). Moreover, verdicts end with a statement about the degree of truthfulness of the related claim, in a form similar to \textit{``We rate the claim \texttt{[TRUTH LABEL]}''}. The main justifications are presented in the body of the verdict. Examples are provided in Table \ref{tab:ex_claim_verdict}.


\begin{table}[h!]
\centering
\small
\begin{tabular}{p{0.95\linewidth}} 
     \toprule 
     \textbf{\texttt{claim}} : Clinton says ``Hate crimes against American Muslims and mosques have tripled after Paris and San Bernardino.''\\
     \midrule 
     \textbf{\texttt{verdict}} : Clinton said that ``\textbf{hate crimes against American Muslims and mosques have tripled after Paris and San Bernardino''.}Calculations by the director of an academic center found that the number did triple after those attacks. But it's worth noting that his data does not show whether or not they remained at that elevated level, or for how long -- something that would be a reasonable interpretation of what Clinton said. The statement is accurate but needs clarification or additional information, so \textbf{we rate it Mostly True}.\\
     \midrule
     \midrule
     \textbf{\texttt{claim}} : Trump says ``I released the most extensive financial review of anybody in the history of politics. ...You don't learn much in a tax return.''\\
     \midrule 
     \textbf{\texttt{verdict}} : Trump said that he has \textbf{``released the most extensive financial review of anybody in the history of politics. ... You don't learn much in a tax return. ''} Trump did release an extensive (and legally required) document detailing his personal financial holdings. However, experts consider that a red herring. Unlike all presidential nominees since 1980, Trump has not released his tax returns, which experts say would offer valuable details on his effective tax rate, the types of taxes he paid, and how much he gave to charity, as well as a more detailed picture of his income-producing assets. Trump's statement is inaccurate. \textbf{We rate it False.}\\
    \bottomrule
\end{tabular}
\caption{Examples from L$_{++}$ : the claim is mostly present within the first sentence, and a truthfulness statement is reported at the end of the verdict.}
\label{tab:ex_claim_verdict}
\end{table}


\section{Fine-tuning Details}
\label{App_fine-tuning}
For the fine-tuning, each model underwent 5 epochs of training with a batch size equal to 4 and a seed set at 2022. To this end, Huggingface Trainer has been employed, keeping its default hyperparameter settings, with the exception of the Learning Rate values and the optimisation method. The Adafactor stochastic optimisation method \citep{shazeer2018adafactor} has been used throughout the whole training phase. Learning Rates values were set as follows: T5 3e-5, Peg$_{xsum}$ 5e-05, Peg$_{cnn}$ 3e-05, dBart 1e-05. For fine-tuning the models, we employed a single GPU, either a Tesla V100 or a Quadro RTX A5000. The checkpoint with minimum \textit{evaluation loss} was used for testing.

\section{Extractive Approach Results Details}
\label{appendix_results}

The first set of experiments tested three main text reduction methodologies: text truncation, LexRank, and SBERT. In order to assess the informativeness of the summaries, the generated extractive output was compared to the gold verdicts through ROUGE metrics. For each methodology, two main configurations have been taken into account: top and bottom (or head and tail for text truncation). While in Table \ref{tab:extractive_sum} we just reported the best configuration (head/top), in Table \ref{tab:extractive_sum_complete} we report the complete results for extractive summarization, which includes the bottom configuration for comparison purposes. These results are confirmed also when these approaches are used for text reduction before the abstractive step in our pipeline. 


\begin{table}[ht!]
\centering
\small
\begin{tabular}{cllccc}
\toprule
\multicolumn{1}{l}{} & \multicolumn{2}{l}{Extraction Method} & R1    & R2    & RL    \\ \midrule
\multirow{6}{*}{\begin{sideways}FF\end{sideways}}       
                                                   & \multirow{2}{*}{truncation}  & head   & 0.258 & 0.082 & 0.182 \\
                                                   &                              & tail   & 0.216 & 0.047 & 0.149 \\
                                                   & \multirow{2}{*}{LexRank}     & top    & 0.267 & 0.083 & 0.180  \\
                                                   &                              & bottom & 0.219 & 0.050 & 0.153 \\
                                                   & \multirow{2}{*}{SBERT}       & top    & 0.300 & 0.114 & 0.213 \\
                                                   &                              & bottom & 0.178 & 0.030 & 0.132 \\ \midrule

\multirow{6}{*}{\begin{sideways}L$_{++}$\end{sideways}}        
                                                   & \multirow{2}{*}{truncation}  & head   & 0.347 & 0.120 & 0.196 \\
                                                   &                              & tail   & 0.313 & 0.061 & 0.157 \\
                                                   & \multirow{2}{*}{LexRank}     & top    & 0.373 & 0.120 & 0.194  \\
                                                   &                              & bottom & 0.302 & 0.056 & 0.154 \\
                                                   & \multirow{2}{*}{SBERT}       & top    & 0.393 & 0.158 & 0.237 \\
                                                   &                              & bottom & 0.245 & 0.029 & 0.131 \\ \bottomrule
\end{tabular}
\caption{Pure extractive approach results for the head/tail and top/bottom configurations. The number of sentences to be extracted is set to the average number of sentences per verdict in the corresponding datasets (2 for FF and 6 for L++) }
\label{tab:extractive_sum_complete}
\end{table}










  

\end{document}